\documentclass[10pt,twocolumn,letterpaper]{article}

\usepackage{iccv}
\usepackage{times}
\usepackage{epsfig}
\usepackage{graphicx}
\usepackage{dcolumn}
\usepackage{amsmath}
\usepackage{amssymb}
\usepackage{xcolor}
\usepackage{color}
\usepackage{multirow}
\usepackage{color}
\usepackage{float}
\usepackage[caption=false]{subfig}
\usepackage{listings}
\usepackage{colortbl}
\usepackage{enumitem}
\usepackage{booktabs}
\usepackage{marvosym}
\usepackage[accsupp]{axessibility}
\definecolor{mygray}{gray}{0.6}
\definecolor{sgreen}{RGB}{30, 150, 30}
\definecolor{green2}{RGB}{50, 200, 100}

\usepackage[pagebackref=true,breaklinks=true,letterpaper=true,colorlinks,bookmarks=false]{hyperref}

\newcolumntype{x}[1]{>{\centering\arraybackslash}p{#1pt}}

\newlength\savewidth\newcommand\shline{\noalign{\global\savewidth\arrayrulewidth
		\global\arrayrulewidth 1pt}\hline\noalign{\global\arrayrulewidth\savewidth}}
\newcommand{\tablestyle}[2]{\setlength{\tabcolsep}{#1}\renewcommand{\arraystretch}{#2}\centering\footnotesize}
\newcommand\blfootnote[1]{%	
  \begingroup
  \renewcommand\thefootnote{}\footnote{#1}%
  \addtocounter{footnote}{-1}%
  \endgroup
}

\iccvfinalcopy % *** Uncomment this line for the final submission

\def\iccvPaperID{6950} % *** Enter the ICCV Paper ID here

\ificcvfinal\fi
\begin{document}

%%%%%%%%% TITLE
\title{TAM: Temporal Adaptive Module for Video Recognition}
\author{Zhaoyang Liu\textsuperscript{\rm 1, 2}
\quad
Limin Wang\textsuperscript{\rm 1 \Letter}
\quad
Wayne Wu\textsuperscript{\rm 2}
\quad
Chen Qian\textsuperscript{\rm 2}
\quad
Tong Lu\textsuperscript{\rm 1}
\\
\textsuperscript{\rm 1} State Key Lab for Novel Software Technology, Nanjing University, China \\
\textsuperscript{\rm 2} SenseTime Research \\
{\tt\small zyliumy@gmail.com} \ \ {\tt\small lmwang@nju.edu.cn} \ \ {\tt\small \{wuwenyan,qianchen\}@sensetime.com} \ \ {\tt\small lutong@nju.edu.cn} \\
}

\maketitle
\ificcvfinal\thispagestyle{empty}\fi

%%%%%%%%% ABSTRACT
\begin{abstract}
Video data is with complex temporal dynamics due to various factors such as camera motion, speed variation, and different activities. To effectively capture this diverse motion pattern, this paper presents a new temporal adaptive module ({\bf TAM}) to generate video-specific temporal kernels based on its own feature map. TAM proposes a unique two-level adaptive modeling scheme by decoupling the dynamic kernel into a location sensitive importance map and a location invariant aggregation weight. The importance map is learned in a local temporal window to capture short-term information, while the aggregation weight is generated from a global view with a focus on long-term structure. TAM is a modular block and could be integrated into 2D CNNs to yield a powerful video architecture (TANet) with a very small extra computational cost. The extensive experiments on Kinetics-400 and Something-Something datasets demonstrate that our TAM outperforms other temporal modeling methods consistently, and achieves the state-of-the-art performance under the similar complexity. The code is available at \url{ https://github.com/liu-zhy/temporal-adaptive-module}.
\end{abstract}
\blfootnote{ \Letter: Corresponding author.}
%%%%%%%%% BODY TEXT
\section{Introduction}
\label{introduction}

Deep learning has brought great progress for various recognition tasks in image domain, such as image classification~\cite{alexnet,resnet}, object detection~\cite{fasterrcnn}, and instance segmentation~\cite{maskrcnn}. The key to these successes is to devise flexible and efficient architectures that are capable of learning powerful visual representations from large-scale image datasets~\cite{imagenet}. However, deep learning research progress in video understanding is relatively slower, partially due to the high complexity of video data. The core technical problem in video understanding is to design an effective temporal module, that is expected to be able to capture complex temporal structure with high flexibility, while yet to be of low computational consumption for processing high dimensional video data efficiently.
\begin{figure*}[t]
  \centering
  \includegraphics[width=13.5cm]{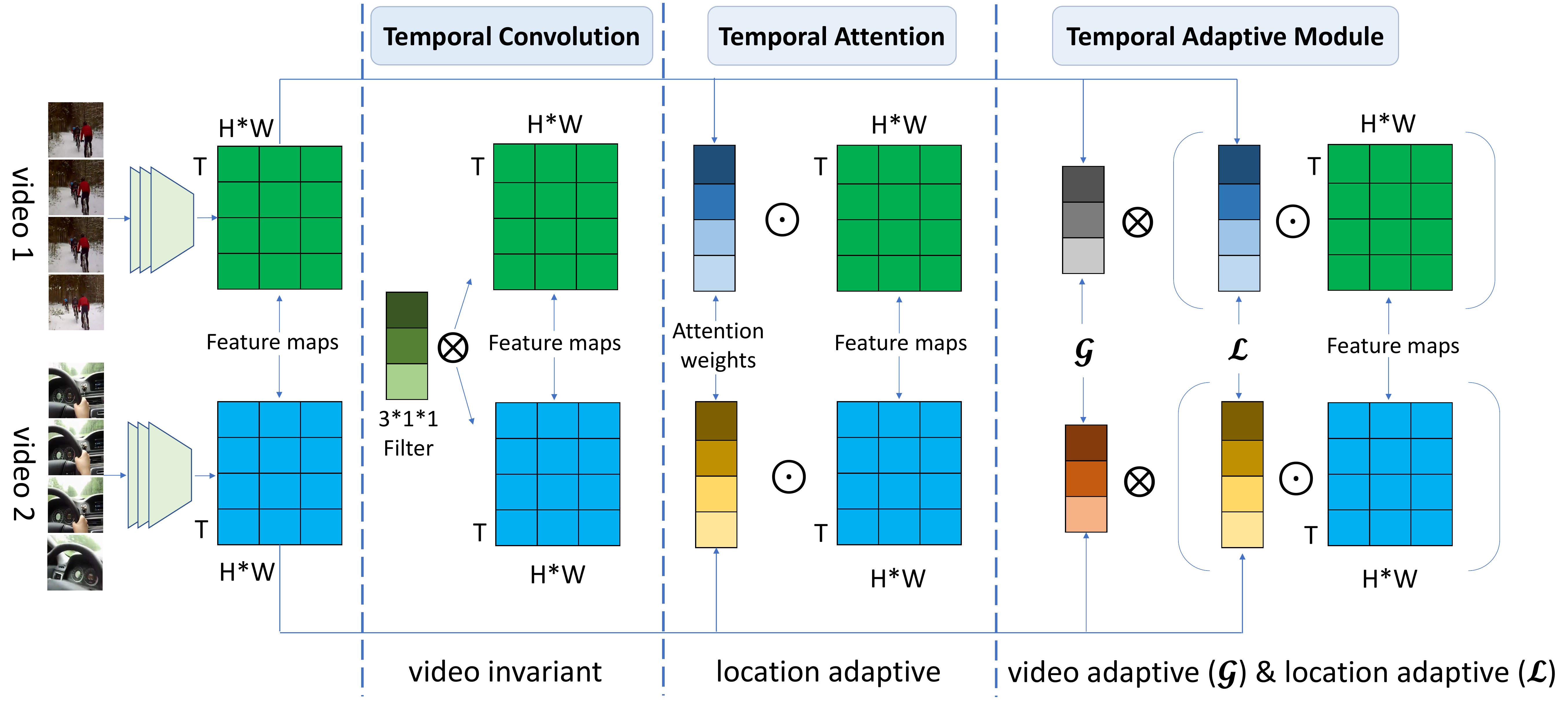}
  \caption{{\bf Temporal module comparisons:} The standard {\em temporal convolution} shares weights among videos and may lack the flexibility to handle video variations due to the diversity of videos. The {\em temporal attention} learns position sensitive weights by assigning varied importance for different time without any temporal interaction, and may ignore the long-range temporal dependencies. Our proposed {\em temporal adaptive module} (TAM) presents a two-level adaptive scheme by learning the local importance weights for location adaptive enhancement and the global kernel weights for video adaptive aggregation. $\odot$
 is attention operation, and $\otimes$ is convolution operation.}
  \label{fig:motivation}
  \vspace{-4mm}
\end{figure*}

3D Convolutional Neural Networks (3D CNNs)~\cite{first_3D,3d_conv} have turned out to be mainstream architectures for video modeling~\cite{I3D,SlowFast,r(2+1)d,p3d}. The 3D convolution is a direct extension over its 2D counterparts and provides a learnable operator for video recognition. However, this simple extension lacks specific consideration about the temporal properties in video data and might as well lead to high computational cost. Therefore, recent methods aim to model video sequences in two different aspects by combining a lightweight temporal module with 2D CNNs to improve efficiency (e.g., TSN~\cite{tsn}, TSM~\cite{tsm}), or designing a dedicated temporal module to better capture temporal relation (e.g., Nonlocal Net~\cite{nonlocal}, ARTNet~\cite{artnet}, STM~\cite{stm}, TDN~\cite{tdn}). However, how to devise a temporal module with both high efficiency and strong flexibility still remains to be an unsolved problem. Consequently, we aim at advancing the current video architectures along this direction.

In this paper, we focus on devising an adaptive module to capture temporal information in a more flexible way. Intuitvely, we observe that video data is with extremely complex dynamics along the temporal dimension due to factors such as camera motion and various speeds. Thus 3D convolutions (temporal convolutions) might lack enough representation power to describe motion diversity by simply employing a fixed number of {\bf video invariant} kernels. To deal with such complex temporal variations in videos, we argue that {\bf adaptive temporal kernels} for each video are effective and as well necessary to describe motion patterns. To this end, as shown in Figure~\ref{fig:motivation}, we present a two-level adaptive modeling scheme to decompose the video specific temporal kernel into a location sensitive {\em importance map} and a location invariant (also video adaptive) {\em aggregation kernel}. This unique design allows the location sensitive importance map to focus on enhancing discriminative temporal information from a local view, and enables the video adaptive aggregation to capture temporal dependencies with a global view of the input video sequence.

\begin{figure}[t]
  \centering
  \includegraphics[width=8cm]{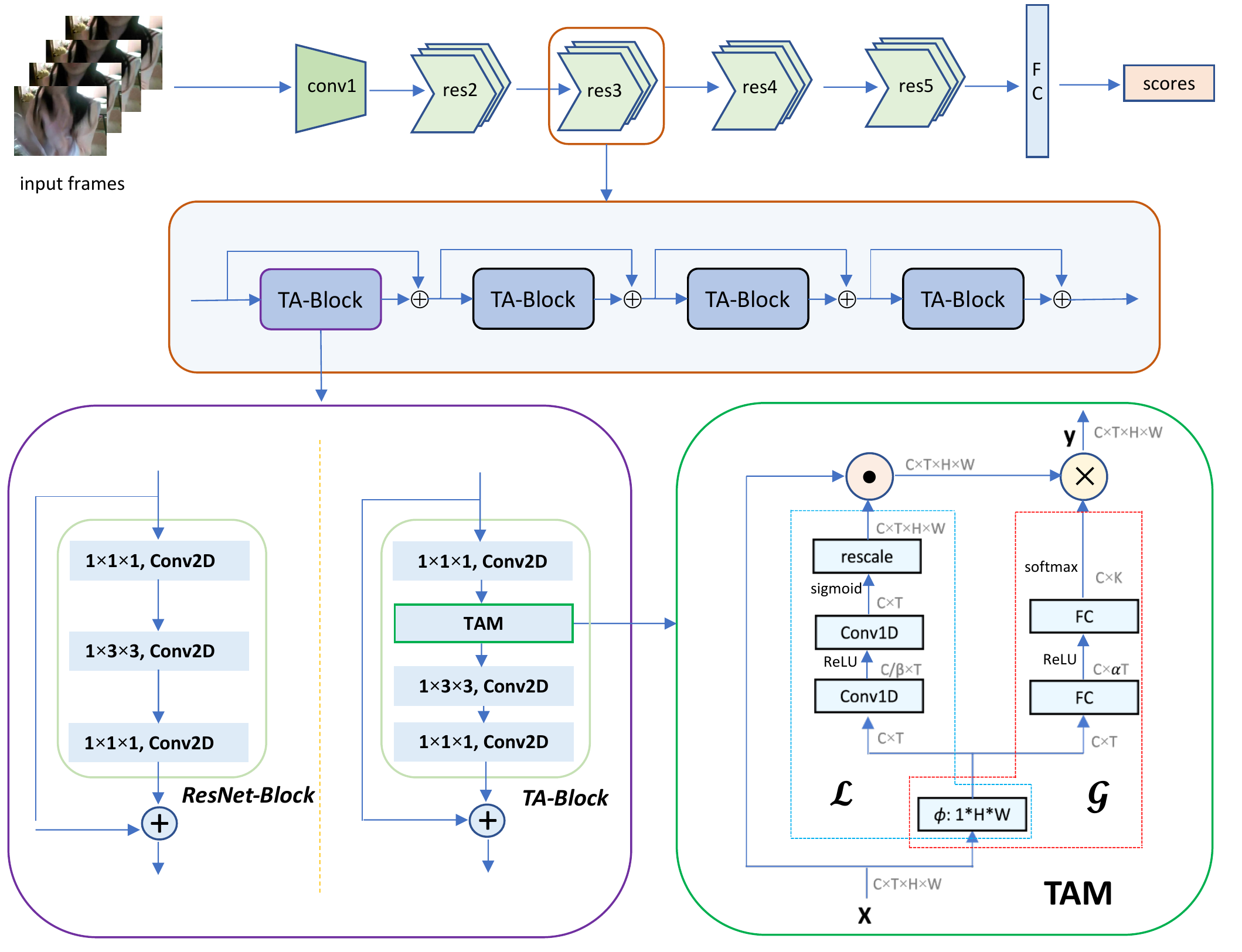}
  \caption{{\bf The overall architecture of TANet:} ResNet-Block \textit{vs}. TA-Block.
  The whole workflow of temporal adaptive module (TAM) in the lower right shows how it works.
  The shape of tensor has noted after each step. $\oplus$ denotes element-wise addition, $\odot$ is element-wise multiplication, and $\otimes$ is convolution operation. The symbols appeared in figure will be explained in Sec.~\ref{sec:tam}.}
  \label{fig:tam}
  \vspace{-4mm}
\end{figure}

Specifically, the design of {\em temporal adaptive module} (TAM) strictly follows two principles: high efficiency and strong flexibility. To ensure our TAM with a low computational cost, we first squeeze the feature map by employing a global spatial pooling, and then establish our TAM in a channel-wise manner to keep the efficiency.
Our TAM is composed of two branches: a local branch ($\mathcal{L}$) and a global branch ($\mathcal{G}$). As shown in Fig.~\ref{fig:tam}, TAM is implemented in an efficient way. The local branch employs temporal convolutions to produce the location sensitive importance maps to enhance the local features, while the global branch uses fully connected layers to produce the location invariant kernel for temporal aggregation. The importance map generated by a local temporal window focuses on short-term motion modeling and the aggregation kernel using a global view pays more attention to the long-term temporal information. Furthermore, our TAM could be flexibly plugged into the existing 2D CNNs to yield an efficient video recognition architecture, termed as TANet.

We verify the proposed TANet on the task of action classification in videos. In particular, we first study the performance of the TANet on the Kinetics-400 dataset, and demonstrate that our TAM is better at capturing temporal information than other several counterparts, such as temporal pooling, temporal convolution, TSM~\cite{tsm}, TEINet~\cite{TEINet}, and Non-local block~\cite{nonlocal}. Our TANet is able to yield a very competitive accuracy with the FLOPs similar to 2D CNNs. We further test our TANet on the motion dominated dataset of Something-Something, where the state-of-the-art performance is achieved.
%------------------------------------------------------------------------
\section{Related Work}
Video understanding is a core topic in the field of computer vision. At the early stage, a lot of traditional methods~\cite{STIPs,HOG3D,action_bank,Extended_SURF} have designed various hand-crafted features to encode the video data, but these methods are too inflexible when generalized to other video tasks. Recently, since the rapid development of video understanding has been much benefited from deep learning methods~\cite{alexnet,vgg,resnet}, especially in video recognition, a series of CNNs-based methods were proposed to learn spatiotemporal representation, and the differences with our method will be clarified later.
Furthermore, our work also relates to dynamic convolution and attention in CNNs.

\noindent \textbf{CNNs-based methods for action recognition.}
Since the deep learning method has been wildly used in the image tasks, there are many attempts~\cite{large_scale_cnn,two_stream,tsn,trn,stnet,tsm, tdn} based on 2D CNNs devoted to modeling the video clips.
In particular, \cite{tsn} used the frames sparsely sampled from the whole video to learn the long-range information by aggregating scores after the last fully-connected layer. \cite{tsm} shifted the channels along the temporal dimension in an efficient way, which yields a good performance with 2D CNNs.
By a simple extension from spatial domain to spatiotemporal domain, 3D convolution~\cite{first_3D,3d_conv} was proposed to capture the motion information encoded in video clips. Due to the release of large-scale Kinetics dataset~\cite{kinetics400}, 3D
CNNs~\cite{I3D} were wildly used in action recognition.
Its variants~\cite{p3d,r(2+1)d,s3d} decomposed the 3D convolution into a spatial 2D convolution and a temporal 1D convolution to learn the spatiotemporal features. And
\cite{SlowFast} designed a network with dual paths to learn the spatiotemporal features and achieved a promising accuracy in video understanding.

The methods aforementioned all share a common insight that they are video invariant and ignore the inherent temporal diversities in videos. As opposed to these methods, we design a two-level adaptive modeling scheme by decomposing the video specific operation into a location sensitive excitation and a location invariant convolution with adaptive kernel for each video clip.

\noindent \textbf{Attention in action recognition.}
The local branch in TAM mostly relates to SENet~\cite{senet}. But the SENet learned modulation weights for each channel of feature maps.
Several methods~\cite{TEINet,STC} also resorted to the attention to learn more discriminative features in videos.
Different from these methods, the local branch keeps the temporal information to learn the location sensitive importances.
\cite{nonlocal} designed a non-local block which can be seen as self-attention to capture long-range dependencies. Our TANet captures the long-range dependencies by simply stacking more TAM, and keep the efficiency of networks.

\noindent \textbf{Dynamic convolutions.}
\cite{dfn} first proposed the dynamic filters on the tasks of video and stereo prediction, and designed a convolutional encoder-decoder as filter-generating network. Several works~\cite{condconv,dc} in image tasks attempted to generate aggregation weights for a set of convolutional kernels, and then produce a dynamic kernel. Our motivation is different from these methods. We aim to use this temporal adaptive module to deal with temporal variations in videos. Specifically, we design an efficient form to implement this temporal dynamic kernel based on input feature maps, which is critical for understanding the video content.

%------------------------------------------------------------------------
\section{Method}
\label{sec:method}
\subsection{The Overview of Temporal Adaptive Module}
\label{sec:tam}
As we discussed in Sec.\ref{introduction}, video data typically exhibit the complex temporal dynamics caused by many factors such as camera motion and speed variations. Therefore, we aim to tackle this issue by introducing a temporal adaptive module (TAM) with video specific kernels, unlike the sharing convolutional kernel in 3D CNNs.
Our TAM could be easily integrated into the existing 2D CNNs (e.g., ResNet) to yield a video network architecture, as shown in Figure~\ref{fig:tam}. We will give an overview of TAM and then describe its technical details.

Formally, let $X\in \mathbb{R}^{C\times T\times H\times W}$ denote the feature maps for a video clip, where $C$ represents the number of channels, and $T, H, W$ are its spatiotemporal dimensions. For efficiency, TAM only focuses on temporal modeling and the spatial pattern is expected to captured by 2D convolutions. Therefore, we first employ a global spatial average pooling to squeeze the feature map as follows:
\begin{equation}
\begin{aligned}
\hat{X}_{c,t} = \phi(X)_{c,t} = \frac{1}{H \times W} \sum_{i,j} X_{c,t,j,i},
\end{aligned}
\label{equ:pooling}
\end{equation}
where $c, t, j, i$ is the index of different dimensions (in channel, time, height and width), and $\hat{X}\in \mathbb{R}^{C\times T}$ aggregates the spatial information of $X$. For simplicity, we here use $\phi$ to denote the function that aggregates the spatial information. The proposed temporal adaptive module (TAM) is established based on this squeezed 1D temporal signal with high efficiency.

Our TAM is composed of two branches: a local branch $\mathcal{L}$ and a global branch $\mathcal{G}$, which aims to learn a location sensitive importance map to enhance discriminative features and then produces the location invariant weights to adaptively aggregate temporal information in a convolutional manner. More specifically, the TAM is formulated as follows:
\begin{equation}
\begin{aligned}
Y = \mathcal{G}(X) \otimes (\mathcal{L}(X) \odot X),
\end{aligned}
\label{equ:arch}
\end{equation}
where $\otimes$ denotes convolution operation and $\odot$ is element-wise multiplication.
It is worth noting that these two branches focus on different aspects of temporal information, where the local branch tries to capture the short term information to attend important features by using a temporal convolution, while the global branch aims to incorporate long-range temporal structure to guide adaptive temporal aggregation with fully connected layers. Disentangling kernel learning procedures into local and global branches turns out to be an effective way in experiments. These two branches will be introduced in the following sections.

\subsection{Local Branch in TAM}
As discussed above, the local branch is location sensitive and aims to leverage short-term temporal dynamics to perform video specific operation.
Given that the short-term information varies slowly along the temporal dimension, it is thus required to learn a location sensitive importance map to discriminate the local temporal semantics.

As shown in Figure~\ref{fig:tam}, the local branch is built by a sequence of temporal convolutional layers with \textit{ReLU} non-linearity.
Since the goal of local branch is to capture short term information, we set the kernel size $K$ as 3 to learn importance map solely based on a local temporal window. To control the model complexity, the first $\mathrm{Conv1D}$ followed by BN~\cite{bn} reduces the number of channels from $C$ to $\frac{C}{\beta}$.  Then, the second $\mathrm{Conv1D}$ with a sigmoid activation yields the importance weights $V \in \mathbb{R}^{C \times T}$ which are sensitive to temporal location. Finally, the temporal excitation is formulated as follows:
\begin{equation}
\begin{aligned}
Z  = \mathrm{F_{rescale}}(V) \odot X = \mathcal{L}(X) \odot X,
\end{aligned}
\label{equ:excitation}
\end{equation}
where $\odot$ denotes the element-wise multiplication and $Z \in \mathbb{R}^{C\times T \times H \times W}$. To match size of $X$, $\mathrm{F_{rescale}}(V)$ rescales the $V$ to $\hat{V}\in \mathbb{R}^{C\times T \times H \times W}$ by replicating in spatial dimension.

\subsection{Global Branch in TAM}
The global branch is location invariant and focuses on generating an adaptive kernel based on long-term temporal information. It incorporates global context information and learns to produce the location invariant and also video adaptive convolution kernel for dynamic aggregation.

\noindent \textbf{Learning the Adaptive Kernels.}
We here opt to generate the dynamic kernel for each video clip and aggregate temporal information in a convolutional manner.
To simply this procedure and as well as preserve high efficiency, The adaptive convolution will be applied in a channel-wise manner. In this sense, the learned adaptive kernel is expected to only model the temporal relations without taking channel correlation into account. Thus, our TAM would not change the number of channels of input feature maps, and the learned adaptive kernel convolves the input feature maps in a channel-wise manner. More formally, for the $c^{th}$ channel, the adaptive kernel is learned as follows:
\begin{equation}
\begin{aligned}
\Theta_c = \mathcal{G}(X)_c = \mathrm{softmax}(\mathcal{F}(\mathbf{W_2}, \delta (\mathcal{F}(\mathbf{W_1}, \phi(X)_c)))),
\end{aligned}
\label{equ:learning_kernel}
\end{equation}
where $\Theta_c \in \mathcal{R}^K$ is generated adaptive kernel (aggregation weights) for $c^{th}$ channel, $K$ is the adaptive kernel size, $\delta$ denotes the activation function \textit{ReLU}. The adaptive kernel is also learned based on the squeezed feature map $\hat{X}_c \in \mathbb{R}^T$ without taking the spatial structure into account for modeling efficiency. But different with the local branch, we use fully connected ($fc$) layers $\mathcal{F}$ to learn the adaptive kernel by leveraging long-term information. The learned adaptive kernel with the global receptive field, thus could aggregate temporal features guided by the global context. To increase the modeling capabilities of the global branch, we stack two $fc$ layers and the learned kernel is normalized with a softmax function to yield a positive aggregation weight. The learned aggregation weights $\Theta = \{\Theta_1, \Theta_2, ..., \Theta_C\}$ will be employed to perform video adaptive convolution.

\noindent \textbf{Temporal Adaptive Aggregation.}
Before introducing the adaptive aggregation, we can look back on how a vanilla temporal convolution aggregates the spatio-temporal visual information:
\begin{equation}
\begin{aligned}
Y  = W \otimes X,
\end{aligned}
\label{equ:normal_conv}
\end{equation}
Where $W$ is the weights of convolution kernel and has no concern with input video samples in inference. We argue this fashion ignores the temporal dynamics in videos, and thus propose a video adaptive aggregation:
\begin{equation}
\begin{aligned}
Y  = \mathcal{G}(X) \otimes X,
\end{aligned}
\label{equ:dynamic_conv}
\end{equation}
where $\mathcal{G}$ can be seen as a kernel generator function. The kernel generated by $\mathcal{G}$ can perform adaptive convolution but is shared cross temporal dimension and still location invariant. To address this issue, the local branch produces $Z$ with location sensitive importance map. The whole procedures can be expressed as follows:
\begin{equation}
\begin{aligned}
Y_{c,t,j,i}  = \mathcal{G}(X) \otimes Z =\Theta \otimes Z = \sum_{k} \Theta_{c, k} \cdot Z_{c,t+k,j,i},
\end{aligned}
\label{equ:fusion}
\end{equation}
where $\cdot$ denotes the scalar multiplication and $Y$ is the output feature maps ($Y \in \mathbb{R}^{C\times T \times H \times W}$).

In summary, TAM presents an adaptive module with a unique  aggregation scheme, where the location sensitive excitation and location invariant aggregation all derive from input features, but focus on capturing different structures (\ie, short-term and long-term temporal structure).

\subsection{Exemplar: TANet}
We here intend to describe how to instantiate the TANet. Temporal adaptive module can endow the existing 2D CNNs with a strong ability to model different temporal structures in video clips. In practice, TAM only causes limited computing overhead, but obviously improves the performance on different types of datasets.

ResNets~\cite{resnet} are employed as backbones to verify the effectiveness of TAM.
As illustrated in Fig.~\ref{fig:tam}, the TAM is embedded into ResNet-Block after the first Conv2D, which easily turns the vanilla ResNet-Block into TA-Block. This fashion will not excessively alter the topology of networks and can reuse the weights of ResNet-Block.
Supposing we sample T frames as an input clip, the scores of T frames after $fc$ will be aggregated by average pooling to yield the clip-level scores. No temporal downsampling is performed before $fc$ layer.
The extensive experiments are conducted in Sec.~\ref{sec:experiments} to demonstrate the flexibility and efficacy of TANet.

\medskip
\noindent \textbf {Discussions.}
We notice that the structure of local branch is similar to the SENet~\cite{senet} and STC~\cite{STC}. The first obvious difference is the local branch does not squeeze the temporal dimension. We thus use temporal 1D convolution, instead of $fc$ layer, as a basic layer. Two-layer design only seeks to make a trade-off between non-linear fitting capability and model complexity.
The local branch provides the location sensitive information, and thereby addresses the issue that the global branch is insensitive to temporal location.

TSN~\cite{tsn} and TSM~\cite{tsm} only aggregate the temporal features with a fixed scheme, but TAM can yield the video specific weights to adaptively aggregate the temporal features in different stages. In extreme cases, our global branch in TAM can degenerate into TSN when dynamic kernel weights $\Theta$ is learned to equal to $[0, 1, 0]$. From another perspective, if the kernel weights $\Theta$ is set to $[1, 0, 0]$ or $[0, 0 ,1]$, global branch can be turned into TSM. It seems that our TAM theoretically provides a more general and flexible form to model the video data.

When it refers to 3D convolution~\cite{first_3D}, all input samples share the same convolution kernel without being aware of the temporal diversities in videos as well.
In addition, our global branch essentially performs a video adaptive convolution whose filter has size $1\times k\times 1\times 1$, while each filter in a normal 3D convolution has size $C \times k \times k \times k$, where $C$ is the number of channels and k denotes the receptive field. Thus our method is more efficient than 3D CNNs. Unlike some current dynamic convolution~\cite{dc,condconv}, TAM is more flexible, and can directly generate the kernel weights to perform video adaptive convolution.

\section{Experiments}
\label{sec:experiments}

\subsection{Datasets}
Our experiments are conducted on three large scale datasets, namely, Kinetics-400~\cite{kinetics400} and Something-Something (Sth-Sth) V1\&V2~\cite{sth}.
Kinetics-400 contains $\sim$300k video clips with 400 human action categories. The trimmed videos in Kinetics-400 are around 10s.
We train the models on the training set ($\sim$240k video clips), and test models on the validation set ($\sim$20k video clips). The Sth-Sth datasets focus on fine-grained and motion-dominated action, which contains pre-defined basic actions involving different interacting objects. The Sth-Sth V1 comprises $\sim$86k video clips in the training set and $\sim$12k video clips in the validation set. Sth-Sth V2 is an updated version of Sth-Sth V1, which contains $\sim$169k video clips in the training set and $\sim$25k video clips in the validation set. They both have 174 action categories.

\subsection{Implementation Details}
\label{sec:implementation}
\noindent\textbf{Training.}
In our experiments, we train the models with 8 and 16 frames as inputs.
On Kinetics-400, following the practice in \cite{nonlocal}, the frames are sampled from 64 consecutive frames in the video. On Sth-Sth V1\&V2, the uniform sampling strategy in TSN~\cite{tsn} is employed to train TANet.
We first resize the shorter side of frames to $256$, and apply the multi-scale cropping and randomly horizontal flipping as data augmentation. The cropped frames are resized to $224\times 224$ for network training.
The batch size is 64. Our models are initialized by ImageNet pre-trained weights to reduce the training time.
Specifically, on the Kinetics-400, the epoch for training is 100. The initial learning rate is set 0.01 and divided by 10 at 50, 75, 90 epochs. We use SGD with a momentum of 0.9 and a weight decay of 1e-4 to train TANet.
On Sth-Sth V1\&V2, we train models with 50 epochs. The learning rate starts at 0.01 and is divided by 10 at 30, 40, 45 epoch. We use a momentum of 0.9 and a weight decay of 1e-3 to reduce the risk of overfitting.

\medskip
\noindent\textbf{Testing.}
Different inference schemes are applied to fairly compare with other state-of-the-art models.
On kinetics-400, we resize the shorter to 256 and take 3 crops of $256\times 256$ to cover the spatial dimensions. In the temporal dimension, we uniformly sample 10 clips for 8-frame models and 4 clips for 16-frame models. The final video-level prediction is yielded by averaging the scores of all spatio-temporal views.
On Sth-Sth V1, we scale the shorter side of frames to 256 and use center crop of $224\times 224$ for evaluation.
On Sth-Sth V2, we employ a similar evaluation protocol to Kinetics, but only uniformly sample 2 clips, and also present the accuracy with a single clip using center crop.

\begin{table*}[t]\centering
		\captionsetup[subfloat]{captionskip=2pt}
		\captionsetup[subffloat]{justification=centering}
		% subfloat ############
		\subfloat[\textbf{Study on parameter choices of $\alpha$ and $\beta$}. The results have revealed an excellent advantage that TAM is not so insensitive to these hyper-parameters.
		\label{tab:r_alpha}]{
		\scalebox{0.9}{
			\tablestyle{4pt}{1.05}
                \begin{tabular}{c|c|c|x{30}x{30}}
                % \toprule[1pt]
                \multicolumn{2}{c|}{Setting} & Frames & Top-1 & Top-5\\
         \shline
         $\alpha$=1 & $\beta$=4 &8 & 75.63\% & 92.10\% \\
         $\alpha$=2 & $\beta$=4  &8 & \textbf{76.28\%} & \textbf{92.60\%} \\
         $\alpha$=4 & $\beta$=4  & 8& 75.72\% & 92.14\% \\
         \shline
         $\alpha$=2 & $\beta$=2 & 8& 75.91\% & 92.38\% \\
         $\alpha$=2 & $\beta$=4 & 8& \textbf{76.28\%} & \textbf{92.60\%} \\
         $\alpha$=2 & $\beta$=8 &8 & 75.63\% & 92.20\% \\
                % \multicolumn{1}{c}{} \\ % space holder
                \end{tabular}}}
                \hspace{9mm}
        % subfloat ############
		\subfloat[\textbf{Study on the temporal receptive fields}. Trying larger temporal receptive fields of $\Theta$ when TANet uses 8 frames and 16 frames as inputs.
		\label{tab:kerner_size}]{
		\scalebox{0.9}{
			\tablestyle{3pt}{1.05}
			 \begin{tabular}{c|c|x{30}x{30}}
                %   \toprule[1pt]
         Kernel & Frames & Top-1 & Top-5\\
         \shline
         K=3 & 8 & 76.28\% & 92.60\%\\
         K=5 & 8 & 75.62\%& 92.14\% \\
         K=3 & 16 & 76.87\% & 92.88\%\\
         K=5 & 16 & 77.19\%& 93.17\% \\
         \multicolumn{4}{c}{}\\
        \end{tabular}}}\hspace{9mm}
        % subfloat ############
		\subfloat[\textbf{Exploring where to insert TAM.} we insert TAM into TA-Block in difference position to study its impact.
		\label{tab:position} ]{
		\scalebox{0.9}{
			\tablestyle{2pt}{1.05}
			\begin{tabular}{x{30}|x{30}|x{30}x{30}}
         Model  & Frames & Top-1 & Top-5\\
         \shline
         TANet-a & 8& 75.95\% & 92.18\% \\
         TANet-b & 8& \textbf{76.28\%} & \textbf{92.60\%} \\
         TANet-c & 8& 75.75\% & 92.13\% \\
         TANet-d & 8& 75.20\% & 91.78\% \\
         \multicolumn{4}{c}{}\\
        %  \multicolumn{4}{c}{}\\
        \end{tabular}}}

        \subfloat[\textbf{Study on the number of TA-Blocks}. The TA-Blocks are cumulatively added into ResNet50 from stage2 to stage5.
        \label{tab:number_of_block}]{
        \scalebox{0.9}{
		\tablestyle{2pt}{1.05}
        \begin{tabular}{c|c|c|x{30}x{30}}
        \multicolumn{4}{c}{} \\ % space holder
        Stages & Frames& Blocks & Top-1 & Top-5\\
         \shline
         res$_5$ & 8 &3 & 74.12\% & 91.45\%\\
         res$_{4-5}$ &8& 9 & 75.15\%& 92.04\% \\
         res$_{3-5}$ & 8& 13 & 75.90\% & 92.22\%\\
         res$_{2-5}$ & 8& 16 & \textbf{76.28\%} & \textbf{92.60\%}\\
        %  \multicolumn{5}{c}{} \\ % space holder
        \end{tabular}}} \hspace{5mm}
          \subfloat[\textbf{Study on the impact on different backbones}. We try to extend the TAM to other backbones. I3D-ResNet-50 takes 32 frame as inputs but other backbones take 8 frame as inputs. The performance shows TAM can easily enjoy the benefits with different backbones.
        \label{tab:other_backbones}]{
        \scalebox{0.9}{
		\tablestyle{2pt}{1.05}
        \begin{tabular}{c|cc|cc|cc|cc|cc}
        % \toprule[1pt]
         \multirow{2}*{Models} &
         \multicolumn{2}{c}{ShuffleNet V2} & \multicolumn{2}{c}{MobileNet V2} & \multicolumn{2}{c}{Inception V3} & \multicolumn{2}{c}{ResNet-50} &
         \multicolumn{2}{c}{I3D-ResNet-50~\cite{deep_analysis}}\\
           & Top-1 & Top-5 & Top-1 & Top-5 &Top-1 & Top-5 & Top-1 & Top-5 & Top-1 & Top-5\\
         \shline
         w/o TAM & 62.1\% & 84.3\% & 64.1\% & 85.6\% & 71.4\% & 89.8\% & 70.2\% & 88.9\% & 76.6\% & - \\
         with TAM & 67.3\% & 87.6\% & 71.6\% & 90.1\% & 75.6\% & 92.0\% & 76.3\% & 92.6\% & 77.2\% & 92.9\% \\
         $\Delta$\textcolor{sgreen}{$Acc.$} & \textcolor{sgreen}{\textit{\textbf{+ 5.2}\%}}  & \textcolor{sgreen}{\textit{\textbf{+ 3.3}\%}} & \textcolor{sgreen}{\textit{\textbf{+ 7.5}\%}}  & \textcolor{sgreen}{\textit{\textbf{+ 4.5}\%}} & \textcolor{sgreen}{\textit{\textbf{+ 4.2}\%}} & \textcolor{sgreen}{\textit{\textbf{+ 2.2}\%}} & \textcolor{sgreen}{\textit{\textbf{+ 6.1}\%}} &
         \textcolor{sgreen}{\textit{\textbf{+ 3.7}\%}} & \textcolor{sgreen}{\textit{\textbf{+ 0.6}\%}} & \textcolor{sgreen}{\textit{\textbf{-}}} \\
		  \multicolumn{11}{c}{} \\ % space holder
          \end{tabular}}} \vspace{2mm}
    \caption{\textbf{Ablation studies on Kinetics-400}. These experiments use ResNet-50 as backbone and take 8-frame as inputs in training. All models share the same inference protocol, i.e., 10 clips $\times$ 3 crops.}
\vspace{-4mm}
\label{tab:exploration}
\end{table*}

\subsection{Ablation Studies}
\label{sec:exploration}
The exploration studies are performed on Kinetics-400 to investigate different aspects of TANet. The ResNet architecture we used is the same with~\cite{resnet}. Our TANet replaces all ResNet-Blocks with TA-Blocks by default.

\medskip
\noindent \textbf{Parameter choices.} We use different combinations of $\alpha$ and $\beta$ to figure out the optimal hyper-parameters in TAM.
The TANet is instantiated as in Fig.~\ref{fig:tam}. TANet with $\alpha=2$ and $\beta=4$ achieves the highest performance shown in Table~\ref{tab:r_alpha}, which will be applied in following experiments.

\medskip
\noindent \textbf{Temporal receptive fields.} We try to increase the temporal receptive fields for learned kernel $\Theta$ in the global branch. From Table~\ref{tab:kerner_size}, it seems the larger $K$ is beneficial to the accuracy when TANet takes more sampled frames as inputs. On the other hand, it even degenerates the performance of TANet when sampling 8 frames. In our following experiments, the $K$ will be set to 3.

\medskip
\noindent \textbf{TAM in the different position.} Table~\ref{tab:position} tries to study the effects of TAM in different position. TANet-a, TANet-b, TANet-c, and TANet-d denote the TAM is inserted before the first convolution, after the first convolution, after the second convolution, and after the last convolution in the block, respectively. These four styles are graphically presented in the {\em supplementary material}.
The style in Fig.~\ref{fig:tam} is TANet-b, which has a slightly better performance than other styles as shown in Table~\ref{tab:position}. The TANet-b will be abbreviated as TANet by default in the following experiments.

\medskip
\noindent \textbf{The number of TA-Blocks.} To make a trade-off between performance and efficiency, we gradually add more TA-Blocks into ResNet. As shown in Table~\ref{tab:number_of_block}, we find that more TA-Blocks contributes to better performance. The res$_{2-5}$ achieves the highest performance and will be used in our experiments.

\medskip
\noindent \textbf{Transferring to other backbones.} Finally we verify the generalization of our proposed module. To this end, we apply the TAM to other well known 2D backbones, like ShuffleNet V2~\cite{ShuffleNetV2}, MobileNet V2~\cite{MobileNetV2}, Inception V3~\cite{inceptionv3} and 3D backbones, like I3D-ResNet-50~\cite{I3D, deep_analysis}, where all models has no temporal downsampling operation before the global average pooling layer. From Table~\ref{tab:other_backbones}, we can observe that the backbone networks equipped with our TAM outperform their C2D and I3D baselines by a large margin, which demonstrates the generalization ability of our proposed module.

\subsection{Comparison with Other Temporal Modules}
\begin{table}[t]
\centering
\vspace{1mm}
\scalebox{0.65}{
\setlength{\tabcolsep}{3mm}{
\begin{tabular}{ccccc}
         \toprule[1pt]
         \multirow{2}*{Models} & FLOPs & \multirow{2}*{Params} & \multirow{2}*{Top-1} & \multirow{2}*{Top-5} \\
         & (of single view)& & &\\
         \shline
         C2D & 42.95G& 24.33M & 70.2\% & 88.9\% \\
         C2D-Pool & 42.95G& 24.33M & 73.1\% & 90.6\% \\
         C2D-TConv & 53.02G& 28.10M & 73.3\% & 90.7\% \\
         C2D-TIM~\cite{TEINet} & 43.06G & 24.37M &  74.7\% & 91.7\% \\
         I3D$_{3 \times 1\times 1}$ & 62.55G & 32.99M &  74.3\% & 91.6\% \\
         \hline
         TSM\textbf{$^\star$}~\cite{tsm} & 42.95G& 24.33M & 74.1\% & 91.2\% \\
         TEINet\textbf{$^\star$}~\cite{TEINet} & 43.01G & 25.11M &  74.9\% & 91.8\% \\

         NL C2D~\cite{nonlocal} & 64.49G & 31.69M& 74.4\% & 91.5\% \\
         \hline
         Global branch & 43.00G & 24.33M & 75.6\% & 91.9\%\\
         Local branch & 43.07G & 25.59M & 73.3\% & 90.7\%\\
         Global branch + SE~\cite{senet} & 43.02G & 24.65M & 75.9\% & 92.1\%\\
         TANet-R & 43.02G & 25.59M & 76.0\% &	92.2\% \\
         TANet & 43.02G & 25.59M & \textbf{76.3}\% & \textbf{92.6}\% \\
         \shline
\end{tabular}
}
}
\vspace{2mm}
\caption{\textbf{Studying on the effectiveness of TAM.} All models use ResNet50 as backbone and take 8 frames with sampling stride 8 as inputs. To be consistent with testing, the FLOPs are calculated with spatial size $256\times256$. \textbf{$\star$} is reported by the author of paper. All methods share the same training setting and inference protocol.}
\label{tab:impact_of_two_modules}
\vspace{-5mm}
\end{table}

As a standard temporal operator, we make comparisons between our TAM and other temporal modules.
For fair comparison, all models in this study employ the same frame input  ($8 \times 8$) and backbone (ResNet-50). The inference protocol is to sample $10 \times 3$ crops to report the performance.

\medskip
\noindent \textbf{Baselines.} We first choose several baselines with temporal modules. We begin with the \textbf{2D ConvNet (C2D)}, where we only build 2D ConvNet with ResNet50 and focus on learning the spatial features. In this sense, it operates on each frame independently without any temporal interaction before the global average pooling layer in the end. The second is the \textbf{C2D-Pool.} To endow the 2D network with temporal modeling capacity, C2D-Pool inserts the average pooling layer whose kernel size is $K\times1\times1$ to perform temporal fusion without any temporal downsampling. This is easily implemented by simply replacing all TAMs in network with average pooling layers. The third type is the learnable temporal convolution, whose kernel is shared by all videos.
We first replace each TAM with a standard temporal convolution with randomly initialized weights, termed as {\bf C2D-TConv}. In addition, we replace the standard temporal convolution with the channel-wised temporal convolution using TSM~\cite{tsm} initialization to solely aggregate temporal information without relating different channels, termed as {\textbf{C2D+TIM}}~\cite{TEINet}. Finally, we compare with \textbf{Inflated 3D ConvNet (I3D)}, whose operation is also based on temporal convolutions by directly inflating the original 2D convolutions into 3D convolutions. In our implementation, we inflate the first $1\times1$ kernel in ResNet-Block to $3\times1\times1$, which can provide a more fair comparison with our TANet. Following~\cite{nonlocal}, this variant is referred to as I3D$_{3\times1\times1}$. It is worth noting these three types of temporal convolutions share the similar idea of fixed aggregation kernel, but differ in the specific implementation details, which can demonstrate the efficacy of adaptive aggregation in our TAM.

The aforementioned methods share the same temporal modeling scheme with a fixed pooling or convolution. As shown in Table~\ref{tab:impact_of_two_modules}, our TAM yields superior performance to all of them. We observe that C2D obtains the worst performance that is less than TAM by 6.1\%. Surprisingly, the naively-implemented temporal convolution (C2D-TConv) performs similar to temporal pooling (C2D-Pool) (73.3\% vs. 73.1\%), which can partly blame on the randomly initialized weights of temporal convolution that corrupt the ImageNet pre-trained weights. In temporal convolution based models, we find that C2D-TIM obtains the best performance with the smallest number of FLOPs. We analyze that this channel-wise temporal convolution can well keep the feature channel correspondence and thus benefits most from the ImageNet pre-trained models. However, it is still worse than our TAM by 1.6\%.

\vspace{2mm}
% \medskip
\noindent \textbf{Other temporal counterparts.} There are some competitive temporal modules that learn video features  based on C2D, i.e.,  \textbf{TSM}~\cite{tsm}, \textbf{TEINet}~\cite{TEINet}, and \textbf{Non-local C2D (NL C2D)}. We here compare our TAM with these different temporal modules, and the results of TSM and TEINet are directly cited from the original papers, as they share similar numbers of FLOPs to our TAM. The non-local block is a kind of self-attention module, proposed to capture the long-range dependencies in videos. The preferable setting with 5 non-local blocks mentioned in \cite{nonlocal} is under a similar computational budget and thereby employed to compare with our TAM. As seen in Table~\ref{tab:impact_of_two_modules}, our TANet achieves highest accuracy among these temporal modules, outperforming TSM by 2.2\%, TEINet by 1.4\%, and NL C2D by 1.9\%.

\vspace{2mm}
% \medskip
\noindent \textbf{Variants of TAM.} To study the performance of each part in temporal adaptive module, we separately validate the \textbf{Global branch} and \textbf{Local branch}. Furthermore, \textbf{Global branch + SE} uses global branch with SE module~\cite{senet} to compare with TANet. TANet achieves the highest accuracy among these models as well, which proves the efficacy of each part of TAM and as well as the strong complementarity between local branch and global branch. We also reverse the order of local branch and global branch (TANet-R): $Y = \mathcal{L}(X) \odot (\mathcal{G}(X) \otimes X)$. We see that TANet is slightly better than TANet-R.

\begin{table}[t]
\centering
\small
\scalebox{0.75}{
\setlength{\tabcolsep}{0.8mm}{
\begin{tabular}{ccccccc}
\shline
\multirow{2}*{Methods} & \multirow{2}*{Backbones} & \multirow{2}*{Training Input} &\multirow{2}*{GFLOPs} & \multirow{2}*{Top-1} & \multirow{2}*{Top-5} \\
 & & & & & & \\

\shline
TSN~\cite{tsn} & InceptionV3 & 3$\times$224$\times$224  & 3$\times$250 & 72.5\% & 90.2\% \\
ARTNet~\cite{artnet} & ResNet18 & 16 $\times$112$\times$112 & 24$\times$250 & 70.7\% & 89.3\% \\
I3D~\cite{I3D} & InceptionV1 & 64$\times$224$\times$224 & 108$\times$N/A & 72.1\% & 90.3\%\\
R(2+1)D~\cite{r(2+1)d} & ResNet34 & 32$\times$112$\times$112 & 152$\times$10 & 74.3\% & 91.4\%\\
\hline
NL I3D~\cite{nonlocal} & ResNet50 & 128$\times$224$\times$224 & 282$\times$30 & 76.5\% & 92.6\% \\
ip-CSN~\cite{csn} & ResNet50 & 8$\times$224$\times$224 & 1.2$\times$10 & 70.8\% & - \\
TSM~\cite{tsm} & ResNet50 & 16$\times$224$\times$224 & 65$\times$30 & 74.7\% & 91.4\%\\
TEINet~\cite{TEINet} & ResNet50 & 16$\times$224$\times$224 & 86$\times$30 & 76.2\% & 92.5\% \\
bLVNet-TAM~\cite{blv-tam} & bLResNet50 & 48$\times$224$\times$224 & 93$\times$9 & 73.5\% & 91.2\% \\
SlowOnly~\cite{SlowFast} & ResNet50 & 8$\times$224$\times$224 & 42$\times$30 & 74.8\% & 91.6\% \\
SlowFast$_{4\times16}$~\cite{SlowFast} & ResNet50 & (4+32)$\times$224$\times$224 & 36$\times$30 & 75.6\% & 92.1\% \\
SlowFast$_{8\times8}$~\cite{SlowFast} & ResNet50 & (8+32)$\times$224$\times$224 & 66$\times$30 & 77.0\% & 92.6\% \\
I3D\textbf{$^\star$}~\cite{deep_analysis} & ResNet50 & 32 $\times$224$\times$224 & 335 $\times$30 & 76.6\% & - \\
\hline
TANet-50 & ResNet50& 8$\times$224$\times$224 & 43$\times$30 & 76.3\% & 92.6\% \\
TANet-50 & ResNet50& 16$\times$224$\times$224 & 86$\times$12 & 76.9\% & 92.9\% \\
\hline
X3D-XL~\cite{X3D} & - & 16$\times$312$\times$312 & 48$\times$30 & 79.1\% & 93.9\% \\

CorrNet~\cite{CorrNet}& ResNet101 & 32$\times$10$\times$3 &  224$\times$30 & 79.2\% &  - \\

ip-CSN~\cite{csn} & ResNet152 & 32 $\times$224$\times$224 & 83$\times$30 & 79.2\% & 93.8\% \\

SlowFast$_{16\times8}$~\cite{SlowFast} & ResNet101& (16+64)$\times$224$\times$224 & 213$\times$30 & 78.9\% & 93.5\% \\
\hline
TANet-101 & ResNet101& 8$\times$224$\times$224 & 82$\times$30 & 77.1\% & 93.1\% \\
TANet-101 & ResNet101& 16$\times$224$\times$224 & 164$\times$12 & 78.4\% & 93.5\% \\
TANet-152 & ResNet152 & 16$\times$224$\times$224 & 242$\times$12 & {\bf 79.3\%} & {\bf 94.1\%} \\

\shline
\end{tabular}
}
}
\vspace{1mm}
\caption{Comparisons with the state-of-the-art methods on Kinetics-400.
As described in \cite{SlowFast}, the GFLOPs of a single view $\times$ the number of views (temporal clips with spatial crops) represents the model complexity. The GFLOPs is calculated with spatial size $256\times 256$. \textbf{$^\star$} denotes the I3D without temporal downsampling.}
\label{tab:state_of_the_kinetics-400}
\vspace{-3mm}
\end{table}

\subsection{Comparison with the State of the Art}
\noindent \textbf{Comparison on Kinetics-400.} Table~\ref{tab:state_of_the_kinetics-400} shows the state-of-the-art results on Kinetics-400. Our method (TANet) achieves the competitive performance to other models. TANet-50 with 8-frame also outperforms SlowFast~\cite{SlowFast} by 0.7\% when using similar FLOPs per view. The 16-frame TANet only uses 4 clips and 3 crops for evaluation such that it provides higher inference efficiency and more fair comparisons with other models. It is worth noting that our 16-frame TANet-50 is still more accurate than 32-frame NL I3D by 1.4\%. As ip-CSN~\cite{csn} is pretrained on Sports-1M~\cite{large_scale_cnn}, it achieves the promising accuracy with deeper backbone, i.e., ResNet152.
Furthermore, TAM is compatible with the existing video frameworks like SlowFast. Specifically, our TAM is more lightweight than a standard $3\times 1 \times 1$ convolution when taking the same number of frames as inputs, but can yield a better performance. TAM thus can easily replace the $3\times 1 \times 1$ convolution in SlowFast to achieve lower computational costs.  X3D has achieved great success in video recognition. X3D was searched by massive computing resources and can not be easily extended in a new situation. Although our method fails to beat all state-of-the-art methods with deeper networks, TAM as a lightweight operator can enjoy the advantages from more powerful backbones and video frameworks. To sum up, the proposed TANet makes a good practice on adaptively modeling the temporal relations in videos.

\medskip
\noindent \textbf{Comparison on Sth-Sth V1 \& V2.}
As shown in Table~\ref{tab:state_of_the_sthv1}, our method achieves the comparable accuracy comparing with other models on Sth-Sth V1. For fair comparison, Table~\ref{tab:state_of_the_sthv1} only reports the results taking a single clip with a center crop as inputs. TANet$_{En}$ is higher than TSM$_{En}$ equipped with same backbone (Top-1: 50.6\% vs. Top-1: 49.7\%). We also conduct the experiments on Sth-Sth V2. V2 has more video clips than V1, which can further unleash the full capabilities of TANet without suffering the overfitting. Following the common practice in~\cite{tsm}, TANets use 2 clips with 3 crops to evaluate the accuracy.
As shown in Table~\ref{tab:state_of_the_sthv2},
our models have achieved the state-of-art performance on Sth-Sth V2. As a result, the TANet$_{En}$ yields a competitive accuracy compared with the two-stream TSM and TEINet$_{En}$.
The results on Sth-Sth V1 \& V2 have demonstrated that our method is also good at modeling the fine-grained and motion-dominated actions.

\begin{table}[t]\small
\centering
\vspace{1mm}
\scalebox{0.6}{
\setlength{\tabcolsep}{1mm}{
\begin{tabular}{c|c|c|c|c|c|c}
\toprule[1pt]
\multirow{2}*{Methods} & \multirow{2}*{Backbones} & \multirow{2}*{Pre-train} &\multirow{2}*{Frames} & \multirow{2}*{FLOPs} & \multirow{2}*{Top-1} & \multirow{2}*{Top-5} \\
 & & & & & & \\
\shline
TSN-RGB~\cite{tsn}& BNInception&ImgNet & $8f$ &16G & 19.5\%& -\\
TRN-Multiscale~\cite{trn}& BNInception & ImgNet & $8f$ & 33G & 34.4\% & -\\
S3D-G~\cite{s3d} & Inception &ImgNet & $64f$ & 71.38G & 48.2\% & 78.7\% \\
ECO~\cite{eco} & BNIncep+Res18 & K400 & $16f$ & 64G & 41.6\% & -\\
ECO$_{En}$Lite~\cite{eco} & BNIncep+Res18& K400 & $92f$ & 267G & 46.4\% & -\\
\hline
TSN~\cite{tsn}& ResNet50 &ImgNet & $8f$ &33G & 19.7\% & 46.6\%\\
I3D~\cite{i3d_gcn}& ResNet50 & ImgNet+K400& $32f\times2$ & 306G & 41.6\% & 72.2\%\\
NL I3D~\cite{i3d_gcn}& ResNet50 & ImgNet+K400 & $32f\times2$ & 334G & 44.4\% & 76.0\%\\
NL I3D+GCN~\cite{i3d_gcn}& ResNet50+GCN & ImgNet+K400 & $32f\times2$ & 606G & 46.1\% & 76.8\%\\
TSM~\cite{tsm} & ResNet50 & ImgNet & $8f$ & 33G & 45.6\% & 74.2\%\\
TSM~\cite{tsm} & ResNet50 & ImgNet & $16f$& 65G & 47.2\% & 77.1\%\\
TSM$_{En}$~\cite{tsm} & ResNet50 & ImgNet & $8f$+$16f$& 98G & 49.7\% & 78.5\%\\
TAM~\cite{blv-tam} & ResNet50 & ImgNet & $8f$ & - & 46.1\% & - \\
bLVNet-TAM~\cite{blv-tam} & ResNet50 & Sth-Sth V1 & $32f$ & 48G & 48.4\% & 78.8\% \\
GST~\cite{gst} & ResNet50 & ImgNet & $8f$ & 30G & 47.0\% & 76.1\% \\
GST~\cite{gst} & ResNet50 & ImgNet & $16f$ & 59G & 48.6\% & 77.9\% \\
TEINet~\cite{TEINet} & ResNet50 & ImgNet & $8f$ & 33G & 47.4\% & - \\
TEINet~\cite{TEINet} & ResNet50 & ImgNet & $16f$ & 66G & 49.9\% & - \\
TEINet$_{En}$~\cite{TEINet} & ResNet50 & ImgNet & $8f$+$16f$ & 66G & 52.5\% & - \\
\hline
 \hline
 TANet & ResNet50 & ImgNet & $8f$ & 33G & 47.3\% & 75.8\%\\
%\cline{3-6}
 TANet & ResNet50 & ImgNet & $16f$ & 66G & 47.6\% & 77.7\%\\
 TANet$_{En}$ & ResNet50 & ImgNet & $8f$+$16f$& 99G & 50.6\% & 79.3\%\\
\shline
\end{tabular}
}
}
\vspace{0.5mm}
\caption{Comparisons with the state-of-the-art methods on Sth-Sth V1. The models only taking RGB frames as inputs are listed in table.  To be consistent with testing, we use  spatial size 224$\times$224 to compute the FLOPs.}
\label{tab:state_of_the_sthv1}
\vspace{-1mm}
\end{table}

\begin{table}[t]
\centering
\scalebox{0.6}{
 \setlength{\tabcolsep}{1mm}{
\begin{tabular}{c|c|c|c|c|c}
\shline
\multirow{2}*{Methods} & \multirow{2}*{Backbones} & \multirow{2}*{Pre-train} &\multirow{2}*{Frames$\times$clips$\times$crops} & \multirow{2}*{Top-1} & \multirow{2}*{Top-5} \\
 & & & & & \\
\shline
TRN~\cite{trn} &BNInception & ImgNet & $8f\times$2$\times$3 & 48.8\% & 77.6\% \\
TSM~\cite{tsm} & ResNet50 & ImgNet & $8f\times$2$\times$3 & 59.1\% & 85.6\% \\
TSM~\cite{tsm} & ResNet50 & ImgNet & $16f\times$2$\times$3 & 63.4\% & 88.5\% \\
TSM$_{2stream}$~\cite{tsm} & ResNet50 & ImgNet & ($16f$+$16f$)$\times$2$\times$3 & 66.0\% & 90.5\% \\
GST~\cite{gst} & ResNet50 & ImgNet & $8f\times$1$\times$1 & 61.6\% & 87.2\% \\
GST~\cite{gst} & ResNet50 & ImgNet & $16f\times$1$\times$1 & 62.6\% & 87.9\% \\
bLVNet-TAM~\cite{blv-tam} & ResNet50 & Sth-Sth V2 & $32f\times$1$\times$1 & 61.7\% & 88.1\% \\
TEINet~\cite{TEINet} & ResNet50 & ImgNet & $8f\times$1$\times$1 & 61.3\% & -\% \\
TEINet~\cite{TEINet} & ResNet50 & ImgNet & $16f\times$1$\times$1 & 62.1\% & -\% \\
TEINet$_{En}$~\cite{TEINet} & ResNet50 & ImgNet & ($8f$+$16f$)$\times$10$\times$3 & 66.5\% & -\% \\
\hline
\hline
TANet & ResNet50& ImgNet & $8f\times$1$\times$1 & 60.5\% & 86.2\%\\
TANet & ResNet50& ImgNet & $8f\times$2$\times$3 & 62.7\% & 88.0\%\\
TANet & ResNet50& ImgNet & $16f\times$1$\times$1 & 62.5\% & 87.6\%\\
TANet & ResNet50& ImgNet & $16f\times$2$\times$3 & 64.6\% & 89.5\%\\
TANet$_{En}$ & ResNet50 & ImgNet & ($8f$+$16f$)$\times$2$\times$3 & 66.0\% & 90.1\%\\
\shline
\end{tabular}
}
}
\vspace{0.5mm}
\caption{Comparisons with the SOTA on Sth-Sth V2. We here apply the two different inference protocal, \ie, 1 clip $\times$ 1 crop and 2 clip $\times$ 3 crop, to fairly evaluate the TAM with other methods.}
\label{tab:state_of_the_sthv2}
\vspace{-3mm}
\end{table}

\begin{figure}[!ht]
  \centering
  \includegraphics[width=8.5cm]{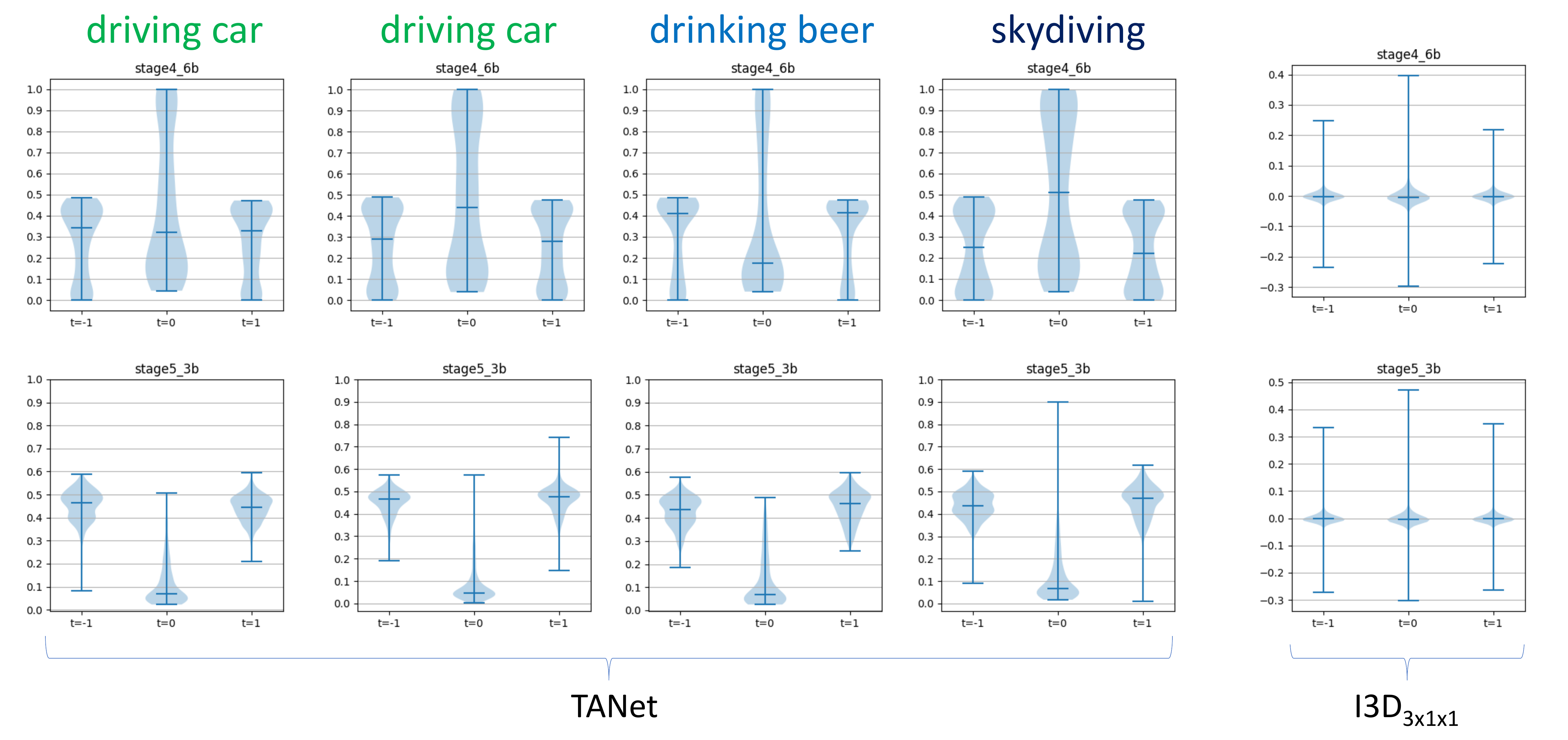}
  \vspace{-3mm}
  \caption{The statistics of kernel weights trained on Kinetics-400, and we plots the distributions in different temporal offsets ($t \in \{-1, 0, 1\}$).
  Each filled area in violinplot represents the entire data range, which marks the minimum, the median and the maximum.
  The first four columns in the left figure are the distributions of learned kernels in TANet.
  In the fifth column, we visualize the filters of $3\times 1\times 1$ kernel in I3D$_{3\times 1\times 1}$ to compare with the TANet.
  }
  \label{fig:vis}
  \vspace{-4mm}
\end{figure}

\subsection{Visualization of Learned Kernels}
\label{sec:visualization}
To better understand the behavior of TANet, we visualize the distribution of kernel $\Theta$ generated by global branch in the last block of stage4 and stage5.
For clear comparison, the kernel weights in I3D$_{3\times 1 \times 1}$ at the same stages are also visualized to find more insights.
As depicted in Fig.~\ref{fig:vis}, we find that the learned kernel $\Theta$ has a different property: the shapes and scales of distribution are more diverse than I3D$_{3\times 1 \times 1}$. Since all video clips share the same kernels in I3D$_{3\times 1 \times 1}$, it causes the kernel weights cluster together tightly. As opposed to temporal convolution, even modeling the same action in different videos, TAM can generate the kernel with slightly different distributions. Taking driving car as an example, the shapes of the distribution shown in Fig.~\ref{fig:vis} are similar to each other but the medians of distributions are not equal.
For different actions like drinking beer and skydiving, the shapes and medians of distributions are greatly different.
Even for different videos of the same action, TAM can learn a different distribution of kernel weights.
Concerning that the motion in different videos may exhibit different patterns, it is necessary to employ an adaptive scheme to model video sequences.

\section{Conclusion}
In this paper, we have presented a generic temporal module, termed as temporal adaptive module (TAM), to capture complex motion patterns in videos and proposed a powerful video architecture (TANet) based on this new temporal module. TAM is able to yield a video-specific kernel with the combination of a local importance map and a global aggregation kernel. This unique design is helpful to capture the complex temporal structure in videos and contributes to more effective and robust temporal modeling.
As demonstrated on the Kinetics-400, the networks equipped with TAM are better than the existing temporal modules in action recognition, which demonstrates the efficacy of our TAM in video temporal modeling. TANet also achieves the state-of-the-art performance on the motion dominated datasets of Sth-Sth V1\&V2.

{\small \paragraph{\bf Acknowledgements.} Thanks to Zhan Tong, Jintao Lin and Yue Zhao for the help. This work is supported by National Natural Science Foundation of China (No. 62076119, No. 61921006), SenseTime Research Fund for Young Scholars, Program for Innovative Talents and Entrepreneur in Jiangsu Province, and Collaborative Innovation Center of Novel Software Technology.}

\appendix
\section{TAM in the different position}
\begin{figure}[t]
  \centering
  \includegraphics[width=8.5cm]{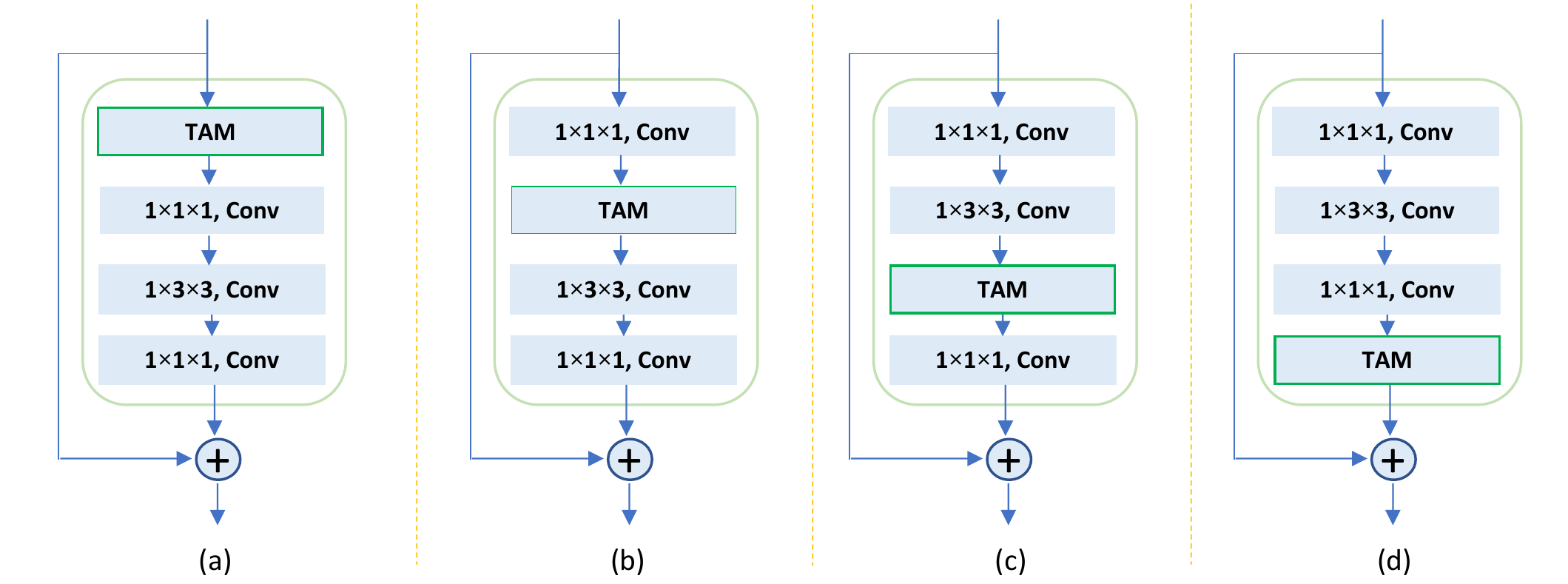}
  \caption{{\bf The four styles of TA-Block.} The (b) is actually the model we used in the main text.}.
  \label{fig:4_style}
%   \vspace{-4mm}
\end{figure}

We here introduce the four different exemplars of TANet. TANet-a, TANet-b, TANet-c, and TANet-d denote the TAM is inserted before the first convolution, after the first convolution, after the second convolution, and after the last convolution in the block, respectively. These four styles are graphically presented in Fig.~\ref{fig:4_style} which were mentioned in main text.

\section{Visualizations of Learned Kernel}
\label{sec:visualization}

In the supplementary material, we are ready to add more visualizations of distribution for {\em importance map $V$} in local branch and {\em video adaptive kernel $\Theta$} in global branch. The 3$\times$1$\times$1 convolution kernels in I3D$_{3\times1\times1}$ are also visualized to study their intentions in inference. To probe into the effects on learning kernels in the different stages, the visualized kernels are chosen in stage4\_6b and stage5\_3b, respectively. Some videos are randomly selected from Kinetics-400 and Sth-Sth V2 to show the diversities in different video datasets.

\begin{figure*}[t]
  \centering
  \includegraphics[width=13cm]{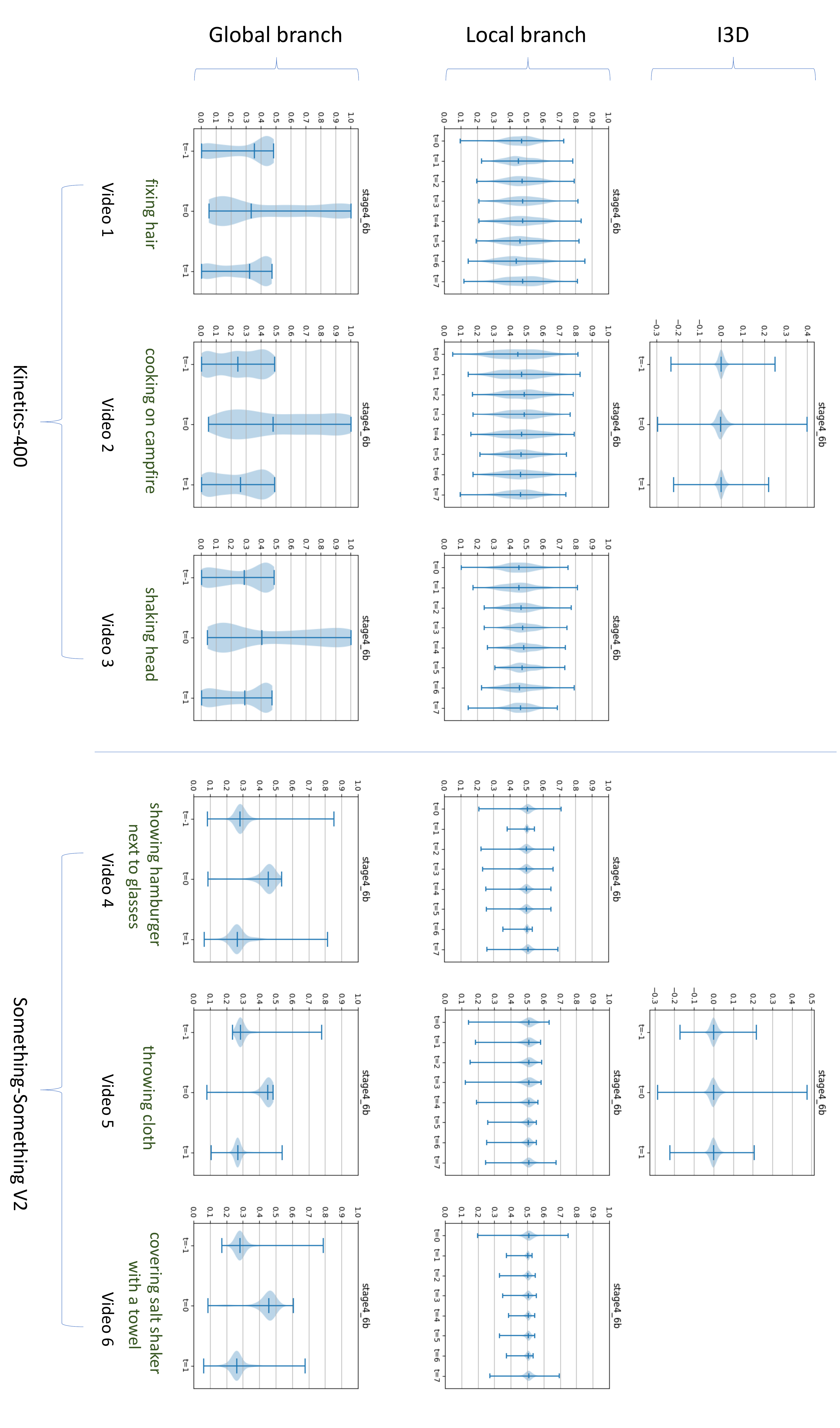}
  \caption{The distribution of learned kernel $V$ and $\Theta$ in the  stage4\_6b.}
  \label{fig:stage4_6b}
\end{figure*}

\begin{figure*}[t]
  \centering
  \includegraphics[width=13cm]{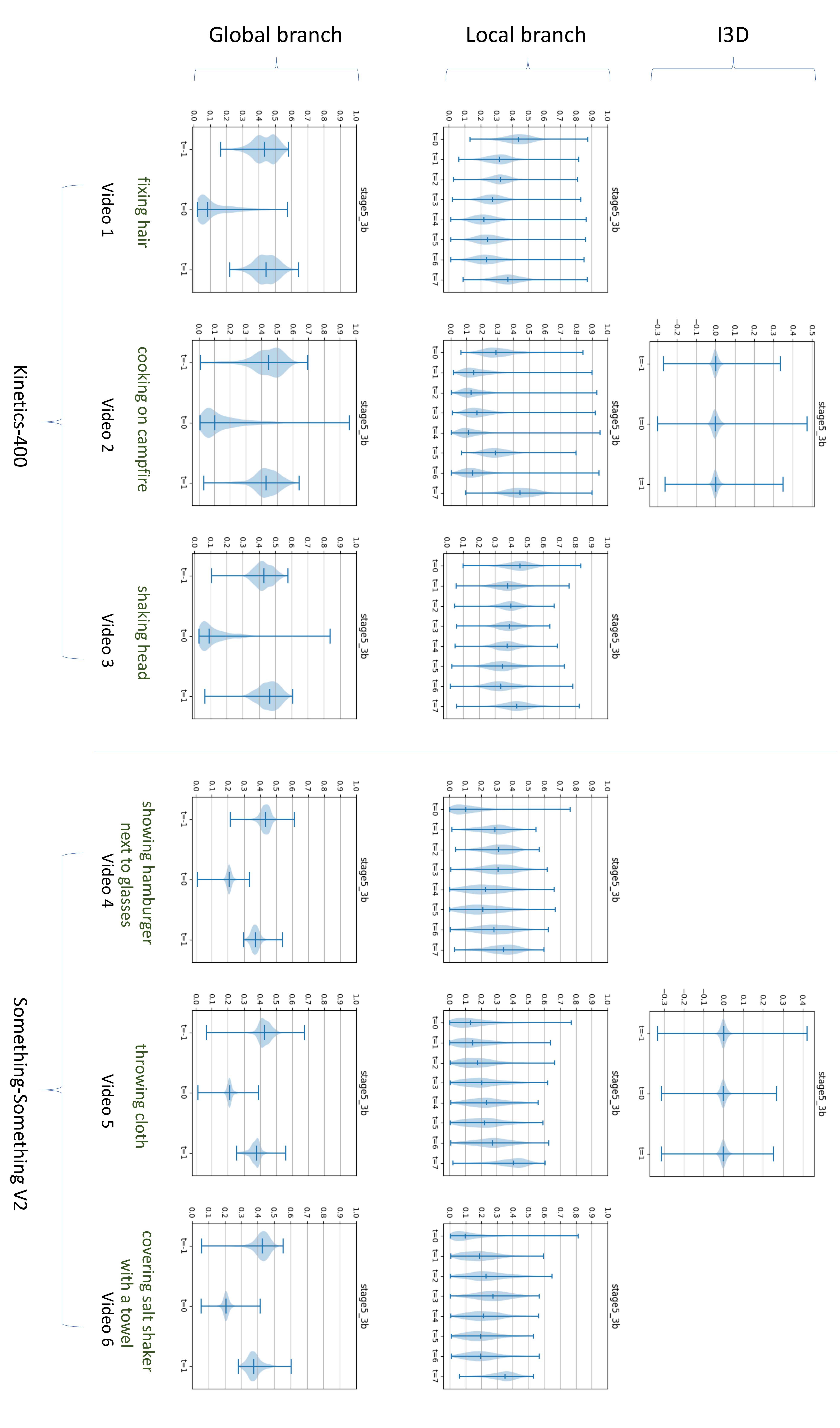}
  \caption{The distribution of learned kernel $V$ and $\Theta$ in the stage5\_3b.}
  \label{fig:stage5_3b}
\end{figure*}

% local branch
Firstly, as depicted in Fig.~\ref{fig:stage4_6b} and Fig.~\ref{fig:stage5_3b}, We can observe that the distributions of importance map $V$ in local branch are smoother than the kernel $\Theta$ in global branch, and local branch pays different attention to each video when modeling the temporal relations.
% global branch
Then, the kernel $\Theta$ in global branch performs the adaptive aggregation to learn the temporal diversities in videos. The visualized kernels in I3D$_{3\times1\times1}$ can make a direct comparison with the kernel $\Theta$, and we find that the distributions of kernel in I3D$_{3\times1\times1}$ are extremely narrow whether on Kinetics-400 or on Sth-Sth V2.
Finally, our learned kernels visualized in figures have exhibited the clear differences between two datasets (Kinetics-400 vs. Sth-Sth V2). This fact is in line with our prior knowledge that there is an obvious domain shift between two datasets.
The Kinetics-400 mainly focuses on appearance and Sth-Sth V2 is a motion dominated dataset. However, this point can not be easily summarized from the kernels in I3D$_{3\times1\times1}$, because the overall distributions of kernels in I3D$_{3\times1\times1}$ on two datasets show minor differences.

Generally, the diversities in our learned kernels have demonstrated that the diversities are indeed existing in videos, and it is reasonable to learn spatio-temporal representation in an adaptive scheme. These findings are again in line with our motivation claimed in the paper.

{\small
\bibliographystyle{ieee_fullname}
\bibliography{egbib}
}

\end{document}